\documentclass{fundam}

\usepackage{url} 
\usepackage[linesnumbered,ruled,lined]{algorithm2e}
\usepackage{graphicx}
\usepackage{tikz}
\usetikzlibrary{automata, positioning, arrows}

\usepackage{booktabs}
\usepackage{multirow}
\usepackage{adjustbox}
\usepackage{afterpage}
\usepackage{enumitem}

\usepackage{amssymb}

\DeclareMathOperator*{\I}{\mathbb{I}}

\begin{document}


\setcounter{page}{29}
\publyear{24}
\papernumber{2176}
\volume{193}
\issue{1}

   \versionForARXIV


\title{Single-sample Versus Case-control Sampling Scheme for Positive Unlabeled Data: the Story of Two Scenarios}

\author{Jan Mielniczuk\thanks{Also affiliated at:  Institute of Computer Science, Polish Academy of Sciences, Warsaw, Poland}
                  \thanks{Address for correspondence:  Faculty of Mathematics and Information Science,
                                Warsaw University of Technology, Koszykowa 75, 00-662 Warsaw, Poland. \newline \newline
                    \vspace*{-6mm}{\scriptsize{Received December 2023; \ accepted July 2024.}}}
\\
    Faculty of Mathematics and Information Science\\
    Warsaw University of Technology\\
    Koszykowa 75, 00-662 Warsaw, Poland \\
    jan.mielniczuk@ipipan.waw.pl
    \and Adam Wawrzeńczyk \\
    Institute of Computer Science \\
    Polish Academy of Sciences\\
    Jana Kazimierza 5, 01-248 Warsaw, Poland \\
    adam.wawrzenczyk@ipipan.waw.pl}

\runninghead{J. Mielniczuk and A. Wawrzeńczyk}{Single-sample Versus Case-control Sampling Schemes for PU Data}

\maketitle

\vspace*{-8mm}
\begin{abstract}
    In the paper we argue that  performance  of the classifiers based on Empirical Risk Minimization (ERM)  for positive unlabeled data, which are  designed for case-control sampling scheme may significantly deteriorate when applied to  a single-sample scenario. We reveal  why  their behavior depends, in all but very specific cases, on the scenario.  Also, we introduce a single-sample case analogue of the popular non-negative risk classifier designed for case-control data and compare its performance with the original proposal. We show that the significant differences occur between them, especially when half or more positive of observations are labeled. The opposite case when ERM minimizer designed for the case-control case is applied for single-sample data is also considered and similar conclusions are drawn. Taking into account difference of scenarios requires a sole, but crucial, change in the definition of the Empirical Risk.

\end{abstract}

\section{Notions and auxiliary results}
\label{section:prelims}

We first introduce basic notations for two scenarios which are commonly encountered when collecting positive unlabeled (PU) data. Let $X$ be a random variable corresponding to feature vector, $Y\in\{-1, 1\}$ be a true class label and $S\in\{-1, 1\}$ an indicator of an example being labeled ($S = 1$) or not ($S = -1$).  Only positive examples ($Y = 1$) from class $P$ can be labeled, i.e. $P(S=1 | X, Y=-1) = 0$. Thus we know that $Y = 1$ when $S = 1$ but when $S = -1$, $Y$ can be either 1 or $-1$. Such situation commonly occurs in medicine when only a part of patients is tested and diagnosed with a certain disease ($S=1$); for the remaining untested patients it is not known whether they are ill ($Y=1$) or not ($Y=-1$). Other areas when such type of restricted observability is encountered include recommendation systems \cite{Schnabel}, survey analysis \cite{BekkerSurvey}, text and image annotation \cite{SVMPenaltyKe} and biology of ecosystems \cite{EM}. The setup is frequently described as missing or weak labels, see e.g. \cite{XMLC}.
In a single-sample scenario (abbreviated to s-s; also called single-training-set or censoring  scenario) we assume that there is some unknown distribution $P_{Y, X, S}$ such that $(Y_i, X_i, S_i), i = 1, \ldots, n$ are independent and identically distributed (iid) random variables drawn from it. Observed data consists of $(X_i, S_i), i = 1,\ldots, n$. Throughout we will assume that the positive observations are selected completely at random for labeling, i.e. labeling does not depend on particular features
\begin{equation}
    \label{eq:SCAR}
    P(S=1 | Y=1, X=x) = P(S=1 | Y=1)
\end{equation}
(SCAR assumption). This in particular implies that the labeled samples are generated from the same distribution as the observations from the positive class i.e. $P_{X | S=1} = P_{X | Y=1}$, see Proposition \ref{eq:SCAR_prop} below. For a treatment of a more general case when the left hand side of (\ref{eq:SCAR}), called a propensity score, depends on $x$ see e.g. \cite{bekker2019ecml}, \cite{LippBoyd2016} and \cite{VAEPUOC}. In contrast, in case-control (c-c) scenario we observe two samples, the first, called labeled class $L$, $X_1, \ldots, X_{n_1}$ pertaining to the positive class $P$ and the second $X_{n_1 + 1},\ldots, X_{n_1 + n_2}$ drawn from a general population being the mixture of distributions $P_X = \pi P_{X | Y=1} + (1 - \pi) P_{X | Y=-1}$ (unlabeled class $U$). Note that c-c scenario is applicable when the samples are drawn from two separate data bases; one pertaining to a general population and the other to patients suffering from a certain disease. On the other hand, when e.g.  a poll is conducted for a randomly chosen group of people who are asked whether they text while driving ($Y=1$) or not ($Y=-1$), this corresponds to s-s scenario  (with affirmative answer resulting in $S=1$).
We refrain from using labeling variable $S$ in the context of c-c data in order to avoid confusion. Note that although for c-c case it would still denote a signed deterministic class indicator (labeled or unlabeled): $S = 2 \times \I\{\text{observation belongs to } L\}-1$, it would be deterministic and not have probabilistic connotation as in s-s case. We remark that the SCAR assumption is automatically satisfied in c-c case if we consider the observations $X_1, \ldots, X_{n_1}$ as labeled: they are all generated from the distribution $P_{X | Y=1}$ and $X_{n_1 + 1}, \ldots, X_{n_1 + n_2}$ are generated from $P_X$.

\begin{figure}[h!]
    \centering
    \includegraphics[width=\textwidth]{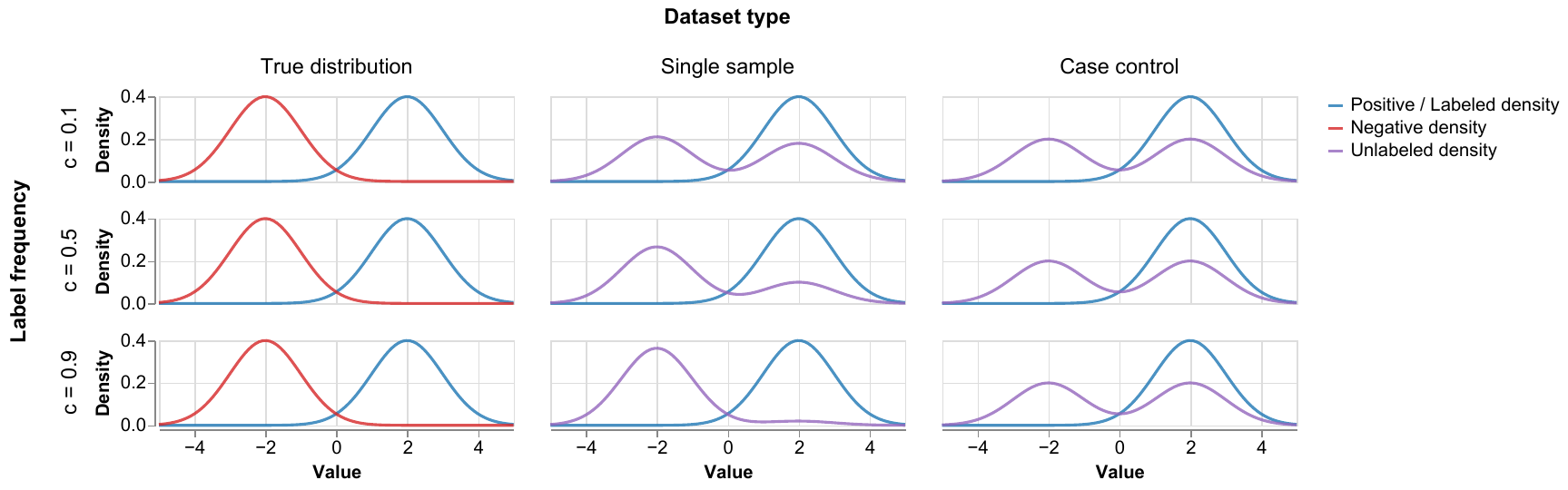}
    \caption{Comparison of labeled and unlabeled class density for s-s and c-c data}
    \label{fig:ss_vs_cc}
\end{figure}

\medskip
Observe that apart from minor differences concerning sample sizes, which are random for single-sample scenario and deterministic for case-control and the important fact that $\pi$ can not be estimated in c-c case; the main difference between the two scenarios lies in a structure of unlabeled sample. In the case-control scenario it corresponds to a general population which is a mixture of $P_{X | Y=1}$ and $P_{X | Y=-1}$ with mixing proportion $\pi = P(Y = 1)$. In contrast, for s-s case it corresponds to the mixture with different mixing proportion
\begin{equation}
    \label{eq:U_dist_ss}
    P_{X | S=-1}
    = \frac{\pi - \pi c}{1 - \pi c} P_{X | Y=1}
    + \frac{1 - \pi}{1 - \pi c} P_{X | Y=-1},
\end{equation}
where $c = P(S=1 | Y=1)$. This easily follows after noticing that out of proportion $\pi$ of positive observations, proportion $\pi c$ is labeled and $\pi -\pi c$ is unlabeled. In particular, we note that it follows from (\ref{eq:U_dist_ss}) that probability that positive element occurs among unlabeled data equals $\pi \times \frac{1 - c}{1 - \pi c}$ and is smaller or equal  $\pi$  being the probability of such occurrence in an original sample. Thus, indeed, distributions of unlabeled samples differ in those cases. In particular note that for $c\approx 1$ unlabeled group in s-s case consists mostly of negative observations in contrast to c-c case when it corresponds to the original mixture. We refer in this context to \cite{BekkerSurvey}, which contains a discussion of both setups. Figure~\ref{fig:ss_vs_cc} shows this behavior for two unit variance normal densities with means $-2$ and 2, respectively, $\pi = 0.5$ and $c=0.1, 0.5$ and $0.9$. Note that whereas for $c = 0.1$ the distributions between unlabeled data in both cases are almost indistinguishable, for $c = 0.9$  there is a striking  difference between unlabeled distributions, the distribution being symmetric and bimodal in c-c case  whereas  in s-s case  the second mode is barely discernible.

There is a legion of papers devoted to inference for PU data, and although they are mostly devoted to c-c scenario (see the comprehensive review \cite{BekkerSurvey}), there are also many approaches which specifically deal with s-s setup, starting from the seminal paper of Elkan and Noto \cite{ElkanNoto}.
However, it seems that understanding of the importance of sampling scenario for behavior of developed classifiers is limited and one can find many examples of careless use of c-c-developed methods in s-s scenario -- sometimes accidentally, but often even after clearly stating the scenario assumptions. This also includes comparing performance of methods designed for a specific scenario with methods for the other scenario, which puts the latter at disadvantage.
The purpose of this paper is to show that this has important consequences and may lead to misleading conclusions especially when establishing a ranking of the  classifiers with respect to some performance metrics. This is analyzed here for a particular case of Empirical Risk Minimizers, which play an important role in PU inference.
We first state a basic fact concerning SCAR scenario.

\begin{proposition}
    \label{eq:SCAR_prop}
    Under SCAR we have for single sample case that $P_{X | S=1} = P_{X | Y=1}$.
\end{proposition}
\begin{proof}
    This easily follows after noting that switching the conditioning we have
    \begin{equation*}
        P(X=x | S=1) = \frac{P(S=1 | X=x) P_X(X = x)}{P(S = 1)}
    \end{equation*}
    and that in view of the fact $S = 1$ implies $Y = 1$ and (\ref{eq:SCAR}) we have
    \begin{equation*}
        \begin{aligned}
              & P(S=1 | X=x) = P(S=1 | Y=1, X=x)P(Y=1 | X=x)                        \\
            = & P(S=1 | Y=1) P(Y=1 | X=x) = \frac{P(S = 1)}{P(Y = 1)} P(Y=1 | X=x).
        \end{aligned}
    \end{equation*}
    Plugging in the formula for $P(S=1 | X=x)$ into the first equation we obtain $P(X=x|Y=1)$ after switching back the conditioning.
\end{proof}

We now derive the form of risk function of classification function $g(X)$ in both scenarios. Let $g(x): \mathbb{R}^p \to \mathbb{R}$ be a classification function with the corresponding classifier being $d(x) = 2 \I\{g(x) \geq 0\} - 1$ (i.e. classifying to positive class $Y = 1$ for $g(x) \geq 0$ and to the negative one in the opposite case). Moreover, $\ell: \mathbb{R} \times \{-1, 1\} \to \mathbb{R}^+$ is a loss function, with $\ell(g(x), y)$ standing for the loss incurred for classification function $g(x)$, when the true class indicator is $y$. Let $R(g)=E_{X, Y} \ell\left(g(X), Y \right)$ be the risk (expected loss) for classification function $g$. From now on, abusing the notion slightly, we consider the losses of the form $\ell(g(x), y) := \ell(g(x) y)$, where $\ell$ is defined now on $\mathbb{R}$.    Thus
\[R(g) =E_{X,Y}\ell(Yg(X)) = \pi E_{X | Y=1} \ell(g(X)) +(1-\pi)E_{X | Y=-1} \ell(-g(X)) .\]

\medskip
We have the following general formula for the risk $R(g)$ which holds regardless the scenario considered:
\begin{equation}
    \label{eq:cc}
    R(g) =E_{X,Y}\ell(Yg(X))
    = \pi E_{X | Y=1} \ell(g(X))
    + E_X \left(\ell(-g(X)) \right)
    - \pi E_{X|Y=1} \left(\ell(-g(X)) \right).
\end{equation}

This immediately follows after noticing that
\begin{equation}
    E_X \left(\ell(-g(X)) \right)
    = \pi E_{X | Y=1} \ell(-g(X))
    + (1 - \pi) E_{X | Y=-1} \ell(-g(X)).
\end{equation}
The formula (\ref{eq:cc}) is directly applicable  in case-control scenario as samples pertaining to $P_X=P_U$ and $P_{X | Y=1}$ are readily available.
In the case of s-s scenario distribution $P_X$ corresponds to distribution of $X$ samples regardless their labels and distribution $P_{X|Y=1}$ equals to
the distribution of the labeled samples $P_{X|S=1}$ in view of Proposition \ref{eq:SCAR_prop}.
Note that $-g(x)$ corresponds to classification function which classifies to the opposite class than $g(x)$.

\medskip
For single-sample scenario we prove a  formula for $R(g)$  which relies solely on $P_{X | S=1}$ and $P_{X | S=-1}$ and which provides additional  insight how procedures designed for c-c case perform in s-s case. This is given by the following Proposition which is a special case of representation of $R(g)$ in \cite{BekkerSurvey}, p. 23, line 3. We note that although (\ref{eq:ss}) is valid in s-s framework, it formally coincides with (\ref{eq:cc}) when $E_{X | S=-1}$
is replaced by $E_X$, $P(S = -1)$ by 1 and $P(Y=1, S=-1)$ by $\pi$.

\begin{proposition}
    In s-s scenario under SCAR $R(g)$ equals
    \begin{equation}
        \label{eq:ss}
        \begin{aligned}
            R(g)
             & = \pi E_{X | S=1} \ell(g(X)) + P(S = -1) E_{X | S=-1} \ell(-g(X)) \\
             & - P(Y=1, S=-1) E_{X | S=1} \ell(-g(X)).
        \end{aligned}
    \end{equation}
\end{proposition}
\begin{proof}
    We show below how (\ref{eq:ss}) follows from another insightful representation (see \cite{NaVAE}), which states that ($E$ stands for $E_{X,Y}$):
    \begin{equation}
        \label{eq:riskI}
        E\left(\ell(Y g(X)) \right)
        = E\left(\ell(S g(X))\right)
        + E\left(
        \left[\ell(g(X)) - \ell(-g(X)) \right]
        \I\{Y=1, S=-1\}
        \right)
    \end{equation}
    holds regardless SCAR condition is valid or not. The above equality can be easily justified by noting that when assigning all unlabeled observations ($S = -1$) to a negative class ($Y = -1$) we commit an error on the set $A=\{Y=1, S=-1\}$. Thus in order to account for this on set $A$ we subtract erroneous part of risk corresponding to $\ell(-g(X))$ and add the correct one $\ell(g(X))$.

\medskip
    In order to prove (\ref{eq:ss}) note that Right-hand Side (RHS) of (\ref{eq:riskI}) can be written in the following form
    \begin{equation}
        \label{eq:riskII}
        \begin{aligned}
            R(g)
             & = E(\ell(g(X)) \I\{S = 1\}) + E(\ell(-g(X))\I\{S = -1\}) \\
             & + E\left(
            \left[\ell(g(X)) - \ell(-g(X)) \right] \I\{Y=1, S=-1\}
            \right)                                                     \\
             & = : I + II + III + IV.
        \end{aligned}
    \end{equation}

    Observe that
    \begin{equation}
        \begin{aligned}
            I + III
             & = P(S = 1) E_{X | S=1} \ell(g(X))           \\
             & + P(Y=1, S=-1) E_{X | Y=1, S=-1} \ell(g(X)) \\
             & = P(S = 1) E_{X | S=1} \ell(g(X))
            + P(Y=1, S=-1) E_{X | Y=1} \ell(g(X))          \\
             & = P(S = 1) E_{X | S=1} \ell(g(X))
            + P(Y=1, S=-1) E_{X | S=1} \ell(g(X))          \\
             & = P(Y = 1) E_{X | S=1} \ell(g(X)),
        \end{aligned}
    \end{equation}
    where the second  equality follows from the conditional independence of $X$ and $S$ given $Y$ under SCAR assumption and the third uses the fact that distributions of $X$ given $Y = 1$ and $X$ given $S = 1$ coincide. Using the same argument we have
    \begin{equation}
        \begin{aligned}
            IV
             & = -P(Y=1, S=-1) E_{X | Y=1, S=-1} \ell(-g(X)) \\
             & = -P(Y=1, S=-1) E_{X | Y=1} \ell(-g(X))       \\
             & = -P(Y=1, S=-1) E_{X | S=1} \ell(-g(X)).
        \end{aligned}
    \end{equation}
    Moreover we have
    \begin{equation}
        II = P(S = -1) E_{X | S=-1} \ell(-g(X)).
    \end{equation}
    From the last three equalities the conclusion follows.
\end{proof}

\begin{remark}
    As a side note we remark that in the case of logistic loss $\ell(s) = \log(1 + e^{-s})$ for which $\ell(s) - \ell(-s) = -s$, equality (\ref{eq:riskI}) for linear classification function $g(x) = \beta^T x$ simplifies to
    \begin{equation*}
        R(g)
        = R(\beta)
        = E\left(\ell(S \beta^T X)\right)
        - P(Y=1, S=-1) \beta^T E_{X | Y=1, S=-1} E X,
    \end{equation*}
    for which correction term $E\left(\ell(Y g(X)) \right) - E\left(\ell(S g(X)) \right)$ is linear function of $\beta$.
\end{remark}

\section{Main result}
We  give now the formal reason why application of ERM classifiers designed for one scenario will fail in the other scenario, save very specific cases. We assume throughout that probability $\pi$ of positive class is known; the assumption being adopted for most of the papers in PU c-c framework and justified by a reasonable assumption that $\pi$ can be precisely approximated from an independent data base. We will focus on Empirical Risk Minimization (ERM) approach, consisting in finding minimizer of an empirical counterpart of the theoretical risk. Consider now a situation when it is applied for s-s data using characteristics of the samples valid in c-c case, namely that labeled samples pertain to $P_{X | Y=1}$ distribution and unlabeled ones are generated from $P_X$. Under SCAR the first assumption is valid as $P_{X | Y=1} = P_{X | S=1}$, whereas the second is not in view of (\ref{eq:U_dist_ss}). Consequently although the first terms in (\ref{eq:cc}) and (\ref{eq:ss}) are equal the second and the third terms in both expressions do not match, suggesting that Empirical Risk Minimization approach for case control situation can not be directly applied to single sample scenario and vice versa. Indeed, closer scrutiny of (\ref{eq:cc}) and (\ref{eq:ss}) yields the following fact.

\begin{proposition} \label{main}\hfill

\noindent  (i) Applying formula (\ref{eq:cc}) for s-s scenario under assumption that $P_U = P_{X | S=-1} = P_X$, is valid only when
    \begin{equation}
        \label{eq:condition}
        E_{X | S=1} \ell(-g(X)) = E_{X | S=-1} \ell(-g(X)),
    \end{equation}
    provided $P(S = 1) > 0$. \medskip\\
    (ii) Conversely, provided $P(S = 1) > 0$, formula (\ref{eq:ss}) for c-c scenario under assumption that $P_U = P_{X | S=-1} = P_X$, is valid only when
    \begin{equation}
        \label{eq:condition2}
        E_{X | S=1} \ell(-g(X)) = E_{X } \ell(-g(X)).
    \end{equation}
\end{proposition}
\begin{proof}
    Application of (\ref{eq:cc}) for s-s scenario means that unlabeled population is treated as the original population, i.e. $P_U = P_{X | S=-1} = P_X$ and thus (\ref{eq:cc}) will take the form
    \begin{equation}
        \label{eq:cc_to_ss}
        \pi E_{X | S=1} \ell(g(X))
        + E_{X | S=-1} \ell(-g(X))
        - \pi E_{X | S=1} \ell(-g(X))
    \end{equation}
    as $E_{X | Y=1} = E_{X | S=1}$. This would be valid formula for $R(g)$ in s-s scenario provided the above expression equals (\ref{eq:ss}), which, taking into account again that under SCAR $P_{X | Y=1} = P_{X | S=1}$, yields
    \begin{equation}
        \begin{aligned}
             & E_{X | S=-1} \ell(-g(X))
            - \pi E_{X | S=1} \ell(-g(X))              \\
             & = P(S = -1) E_{X | S=-1} \ell(-g(X))    \\
             & - P(Y=1, S=-1) E_{X | S=1} \ell(-g(X)).
        \end{aligned}
    \end{equation}
    Thus, using $\pi - P(Y=1, S=-1) = P(S=1)$, we have that the equality above is equivalent to
    \begin{equation}
        P(S = 1) E_{X | S=-1} \ell(-g(X))
        = P(S = 1) E_{X | S=1} \ell(-g(X)),
    \end{equation}
    which yields the conclusion of (i).

    \eject
   \noindent  Proof of (ii) is analogous. Formula (\ref{eq:ss}) under assumption $P_U = P_{X | S=-1} = P_X$ takes the form
    \begin{equation*}
        \pi E_{X | S=1} \ell(g(X))
        + P(S = -1) E_{X} \ell(-g(X))
        - P(Y=1, S=-1) E_{X | S=1} \ell(-g(X))
    \end{equation*}
    and equating it to (\ref{eq:cc}) yields condition (\ref{eq:condition2}).
\end{proof}

Thus it follows that applying Empirical Risk Minimization (ERM) methods suitable for c-c PU data (such as popular $uPU$ and $nnPU$ methods) to s-s data directly, without modifying them, puts them at disadvantage.

Other  methods derived for PU case-control situation will also be affected when applied to single-sample data.  This is due to the fact that they necessarily use the information that unlabeled observations follow the  general distribution. We have focused here on ERM methods as in this case it is possible to formally analyze the impact of scenario on the method; see Proposition \ref{main}.
In the following we compare performance of $nnPU_{cc}$, which optimizes non-negative modification of the empirical version of (\ref{eq:cc}), with its counterpart adapted to s-s data, which will be called $nnPU_{ss}$ in the sequel. We compare their performance in a single-sample case (when $nnPU_{ss}$ should be used) and in a case-control case (when the alternative method should be used). 

\begin{remark}
    We note that both (\ref{eq:condition}) and (\ref{eq:condition2}) are implied by $P_{X | S=1} = P_{X | S=-1}$. However, it is easy to check that due to (\ref{eq:U_dist_ss}) and equality $P_{X | Y=1} = P_{X | S=1}$, this is equivalent to $P_{X | Y=1} = P_{X | Y=-1}$, which makes the original problem void, as distinguishing between classes is clearly impossible in this case.
\end{remark}

\subsection{Empirical risks}

Consider first two  popular estimators of (\ref{eq:cc})  constructed for c-c data.
Note that $R(g)$ in equation~(\ref{eq:cc}) can be split into three components: the  first, corresponding to the labeled sample risk (which we will denote as $R^L$ in the algorithm), the second -- to the risk with respect to  the general  distribution $P_X$ (denoted by $R^D$), as well as PU SCAR correction of the second term ($R^{corr}$). The first and the third component of Eq.~(\ref{eq:cc}) are independent of the scenario. The general distribution  component $E_Xl(-g(X))$  can be consistently  approximated by empirical average over $U$ as  observations in $U$ are distributed according to $P_X$.
 Thus the direct plug-in estimator of (\ref{eq:cc}) is (\cite{uPU})
\begin{equation}
    \label{eq:uPU}
    \begin{aligned}
        \hat R_{uPU_{cc}}(g)
         & =  \frac{\pi}{n_L} \sum_{i: X_i\in L} \ell(g(X_i))
        + \frac{1}{n_U} \sum_{i: X_i\in U} \ell(-g(X))           \\
         & -  \frac{\pi}{n_L} \sum_{i: X_i\in L} \ell (-g(X_i)).
    \end{aligned}
\end{equation}
The minimizer of (\ref{eq:uPU}) is called unbiased PU estimator ($uPU$) and the index 'cc' is added in order to stress that it is derived for c-c data. The nonnegative version ($nnPU$) is obtained when truncating the sum of the two last elements in (\ref{eq:uPU}) at 0. This is motivated by the fact that its  theoretical counterpart  $E_X\ell(-g(X))-\pi E_{X|Y=1}\ell(-g(X))$ is nonnegative.
\eject
\noindent Thus $nnPU_{cc}$ estimator is minimizer of (see \cite{PUSB}):
\begin{equation}
    \label{nnPU}
    \begin{aligned}
        \hat R_{nnPU_{cc}}(g)
         & = \frac{\pi}{n_L} \sum_{i: X_i \in L} \ell(g(X_i))          \\
         & + \max\left(\frac{1}{n_U} \sum_{i: X_i \in U} \ell(-g(X_i))
        - \frac{\pi}{n_L} \sum_{i: X_i \in L} \ell(-g(X_i)), 0 \right),
    \end{aligned}
\end{equation}
We consider now  $nnPU_{ss}$  estimator defined as the minimizer of the risk $R(g)$ for s-s scenario  based on (\ref{eq:cc}).
Note that for this scenario term  $E_Xl(-g(X))$ can be approximated by empirical average over {\it all} observations disregarding their labels and thus we have direct plug-in estimator $\hat{R}_{uPU_{ss}}$
\begin{equation}
    \label{eq:uPUss_alternative}
    \begin{aligned}
        \hat{R}_{uPU_{ss}}
         & = \frac{\pi}{n_{L}} \sum_{X_i \in L} l(g(X_i))                                                 \\
         & + \frac{1}{n} \sum_{X_i\in L \cup U} l(-g(X_i)) -  \frac{\pi}{n_{L}} \sum_{x \in L} l(-g(X_i))
    \end{aligned}
\end{equation}
and its nonnegative version $\hat{R}_{nnPU_{ss}}$:
\begin{equation}
    \label{eq:nnPUss_alternative}
    \begin{aligned}
        \hat{R}_{nnPU_{ss}}
         & = \frac{\pi}{n_{L}} \sum_{X_i \in L_i} l(g(X_i))                                                                      \\
         & + \max\left(\frac{1}{n} \sum_{X_i\in L \cup U} l(-g(X_i)) -  \frac{\pi}{n_{L_i}} \sum_{x \in L} l(-g(X_i)), 0\right).
    \end{aligned}
\end{equation}
The final, labeling-scenario-aware training procedure is described in detail in Algorithm~\ref{algorithm:nnPU-upgrade}. Note that algorithm incorporates the change of the gradient sign when truncation occurs as advocated in the original $nnPU$  algorithm (\cite{nnPU}). We also note that  in order to obtain the version of $nnPU_{ss}$ algorithm it is only necessary to change in the definition of  $\hat{R}_{uPU_{ss}}$, the value of $R^D$ to the average calculated over all data and not over $U$ sample.
\begin{remark}
    We also note that it is possible to obtain  $\hat{R}_{uPU_{ss}}$ based on (\ref{eq:ss}). Namely plug-in version of (\ref{eq:ss})
    has the form
    \begin{equation}
        \label{eq:correctERM}
        \begin{aligned}
             & \frac{\pi}{n_L} \sum_{i: X_i\in L} \ell(g(X_i))
            + \frac{1}{n} \sum_{i: X_i\in U} \ell(-g(X))                                         \\
             & - \left(\pi - \frac{n_L}{n} \right) \frac{1}{n} \sum_{i: X_i\in L} \ell(-g(X_i)),
        \end{aligned}
    \end{equation}
    where $n_L = \left|\{i: S_i = 1\} \right|$ and $n_U = n - n_L$ and $P(Y=1, S=-1)$ is estimated by $\pi - \hat{P}(S = 1) = \pi - \frac{n_L}{n}$.
    Its non-negative version is defined analogously. However, by splitting the sum $\frac{1}{n} \sum_{X_i\in L \cup U} l(-g(X_i))$ into the sums over $L$ and $U$
    it is easy to see that (\ref{eq:correctERM}) equals (\ref{eq:uPUss_alternative}). The  form of the risk function in (\ref{eq:uPUss_alternative}), while less useful from the theoretical point of view, has an important advantage: namely that only a single component differs between (\ref{eq:uPU}) and (\ref{eq:uPUss_alternative}). This allows for easier comparison of those risks over the course of our experiments and is thus used in our implementation of the $nnPU_{ss}$ algorithm.
 \end{remark}

We discuss the reason the negative part of $\hat R_{uPU_{cc}}$ is biased downwards when applied to single sample data. The first and the third terms in (\ref{eq:uPU}) are consistent estimators of their theoretical counterparts, the problem occurs for the second term. Namely, large contribution to the sum $\sum_{i \in U} \ell(-g(X_i))$ pertains to positive elements which are likely to be assigned to positive class by classification function $g(x)$ but will be assigned to negative class by $-g(x)$ thus producing large loss. However, when $uPU_{cc}$ is applied to s-s data the proportion of positive elements among unlabeled data is smaller than in the general population as discussed in the introduction. This results in downward bias of the negative part of ERM. Thus truncation at 0 is more likely to occur here than for c-c scenario. Conversely, when $nnPU_{ss}$ is applied for c-c data the term ${n}^{-1} \sum_{i: S_i = -1} \ell(-g(X_i))$ in (\ref{eq:correctERM}) is larger than for s-s data as the proportion of positive observations among unlabeled observations is larger. Note that the both cases of misuse are not entirely symmetric as erroneous application of $\hat R_{nnPU_{cc}}(g)$ to s-s data results in more likely truncation, whereas for $\hat R_{nnPU_{ss}}(g)$ used for c-c data truncation is less likely.

\section{Numerical experiments}

We consider two sampling scenarios:
\begin{enumerate}[label=(\alph*)]
    \item \textbf{Single-sample scenario}. For a given data set for which $\pi$ is taken as a fraction of positive class in it, we sample $n=1000$ elements randomly without replacement and label them using SCAR scenario with varying $c=P(S=1|Y=1)$. We apply s-s and c-c methods to the sample obtained.
    \item \textbf{Case-control scenario.} In the following $n$ again will stand for the total number of observations. In order to ensure that  the expected fraction of labeled observations for a given $c$ is the same  in both scenarios we pick $c\times\pi\times n$ observations from positive class and $(1-c)\times n$ from the whole data set. Thus $ c\times\pi\times n+ \pi\times(1-c)\times n =\pi\times n $ is an expected number of observations from the positive class in the sample and fraction $c$ of them will be labeled on average. In order to ensure that the chosen sample has size equal to $n$ both sizes should be increased $A=(1-c(1-\pi))^{-1}$ times i.e. size of chosen labeled sample should be $A\times c\times\pi\times n$ and $A\times(1-c)\times n$ for the unlabeled one. Note that both samples are not necessarily disjoint but this should not affect the performance of the rule as empirical risk function is an unbiased estimator of theoretical risk also in this case.
          We stress  that examination of these two sampling scenarios shows that the meaning of some parameters is different in both methods, e.g. for single-sample scenario $c=P(S=1|Y=1)=1$ means that every positive observation will be labeled and thus the sample $(X_i,S_i)_{i=1}^n$ is generated from the distribution $P_{X,Y}$. In case control scenario it means that every observation for positive class is sampled, however unlabeled set will be empty as otherwise (see the calculation above) $c$ will be strictly smaller than 1.
\end{enumerate}

We performed experiments on a collection of 18 diverse datasets from different domains: image, text and tabular classification. Dataset details are shown in table \ref{tab:datasets} -- the test ensemble contains datasets of various sizes and class balances. In order  to focus on risk function differences rather than on tuning particular network architectures, as well as to shorten training time, we used pretrained embeddings for text and image data (\verb+all-MiniLM-L6-v2+ \cite{MiniLM} and \verb+swiftformer-xs+ \cite{SwiftFormer} respectively). For processed data classification, we used 5-layer feed-forward neural network, matching the one used in \cite{nnPU}. Similarly to this paper, we kept the default values of hyper-parameters: $\beta = 0$ and $\gamma = 1$ (see Algorithm \ref{algorithm:nnPU-upgrade} for their description). Based on the each dataset, we synthetically created both single-sample and case-control problem, as described above. We either use a pre-existing train-test split or, in cases where it is not available, split dataset with a 80-20 training-test ratio.

\begin{table}[h!]
\vspace*{-2mm}
    \centering
    \caption{Dataset statistics}
    \scalebox{0.8}{
        \begin{tabular}{lc|ccc}
\toprule
Dataset & Data type & Samples & Features & $\pi$ \\
\midrule
CIFAR & Image & 50000 & 10780 & 0.60 \\
MNIST & Image & 60000 & 10780 & 0.51 \\
FashionMNIST & Image & 60000 & 10780 & 0.50 \\
EuroSAT & Image & 21600 & 10780 & 0.30 \\
Chest X-ray & Image & 4077 & 10780 & 0.73 \\
Snacks & Image & 4838 & 10780 & 0.41 \\
DogFood & Image & 2250 & 10780 & 0.33 \\
Beans & Image & 1034 & 10780 & 0.33 \\
Oxford Pets & Image & 5912 & 10780 & 0.33 \\
20News & Text & 11314 & 384 & 0.56 \\
IMDB & Text & 25000 & 384 & 0.50 \\
HateSpeech & Text & 8561 & 384 & 0.11 \\
SMSSpam & Text & 4459 & 384 & 0.13 \\
PoemSentiment & Text & 892 & 384 & 0.15 \\
Credit & Tabular & 13371 & 10 & 0.50 \\
California & Tabular & 16507 & 8 & 0.50 \\
Wine & Tabular & 2043 & 11 & 0.51 \\
Electricity & Tabular & 30779 & 7 & 0.50 \\
\bottomrule
\end{tabular}

    }
    \label{tab:datasets}
\end{table}

\begin{algorithm}[htbp]
    \caption{Scenario-aware nnPU algorithm}
    \label{algorithm:nnPU-upgrade}
    \KwIn{Positive-unlabeled dataset $X = (L, U)$, $\pi$~--~class prior, $n$~--~number of training items, hyperparameters $\beta$ and $\gamma$ (as described in detail in~\cite{nnPU}).}

    \Repeat{not converged}{
        Split $X$ into $k$ minibatches

        \ForAll{minibatch $M_i = (L_i, U_i)$ in $X$}{
            Calculate labeled risk component $R^{L}$
            \begin{equation*}
                R^{L} = \pi \frac{1}{n_{L_i}} \sum_{x \in L_i} l(g(x)),
            \end{equation*}

            \uIf{nnPU$_{ss}$}{
                Calculate general distribution component $R^{D}$ based on the whole dataset
                \begin{equation*}
                    R^{D} = R^D_{ss} = \frac{1}{n} \sum_{x \in L_i \cup U_i} l(-g(x)),
                \end{equation*}
            }
            \uElseIf{nnPU$_{cc}$}{
                Calculate general distribution component $R^{D}$ based on the unlabeled set
                \begin{equation*}
                    R^{D} = R^D_{cc} = \frac{1}{n_{U_i}} \sum_{x \in U_i} l(-g(x)),
                \end{equation*}
            }
            Calculate PU SCAR correction $R^{corr}$
            \begin{equation*}
                R^{corr} = \pi \frac{1}{n_{L_i}} \sum_{x \in L_i} l(-g(x)),
            \end{equation*}
            \uIf{nonnegative risk component $R^{D} - R^{corr} \leq -\beta$}{
                Perform gradient descent for unbiased risk $R = R^{L} + (R^{D} - R^{corr})$ with step size $\eta$.
            }
            \Else{
                \label{algorithm:nnPU-upgrade:negative-correction}
                Update model parameters using surrogate $R^{surr} = R^{corr} - R^{D}$ with discounted step size $\gamma \eta$.
            }
        }
    }
\end{algorithm}

During testing, multiple label frequency $c$ levels, ranging from $0.1$ to $0.9$, were applied. Each experiment (for a given dataset, scenario, label frequency and method combination) was repeated 10 times with a different random seed. As the paper's aim is to emphasize impact of correct method selection, we focused on comparing $nnPU_{ss}$ and $nnPU_{cc}$ methods and did not include comparisons with any external classifiers. Implementation of Algorithm~\ref{algorithm:nnPU-upgrade} and all experiments' code are publicly available on GitHub\footnote{\url{https://github.com/wawrzenczyka/nnPUss}}.

\subsection{Results}

\begin{table}[htbp]
    \centering
    \scriptsize
    \setlength{\tabcolsep}{3.5pt}
    \renewcommand{\arraystretch}{0.9}
    \caption{Test accuracy, single-sample datasets. $\Delta$ indicates accuracy difference between scenario-appropriate $nnPU_{ss}$ method and ill-specified $nnPU_{cc}$ method.}
    \begin{adjustbox}{center}
        \begin{tabular}{l|c|ccccccccc}
\toprule
	\textbf{c} & \textbf{Model} & \textbf{Beans} & \textbf{CIFAR} & \textbf{Chest X-ray} & \textbf{DogFood} & \textbf{EuroSAT} & \textbf{FashionMNIST} & \textbf{MNIST} & \textbf{Oxford Pets} & \textbf{Snacks} \\
\midrule
	\multirow[c]{3}{*}{0.1} & nnPUcc & 81.88 & 92.69 & 88.93 & 87.69 & 90.26 & 97.16 & 95.16 & 86.80 & 74.96 \\
	 & nnPUss & 79.22 & 91.53 & 89.62 & 86.15 & 87.77 & 95.14 & 93.71 & 83.50 & 75.05 \\
	 & $\Delta$ & -2.66 & -1.16 & 0.69 & -1.55 & -2.49 & -2.02 & -1.45 & -3.29 & 0.09 \\
\midrule
	\multirow[c]{3}{*}{0.3} & nnPUcc & 89.92 & 92.01 & 91.49 & 97.56 & 94.19 & 96.68 & 95.23 & 95.86 & 80.68 \\
	 & nnPUss & 89.06 & 93.80 & 92.71 & 95.21 & 91.71 & 96.96 & 96.77 & 89.76 & 81.34 \\
	 & $\Delta$ & -0.86 & 1.78 & 1.22 & -2.35 & -2.48 & 0.28 & 1.53 & -6.10 & 0.66 \\
\midrule
	\multirow[c]{3}{*}{0.5} & nnPUcc & 91.09 & 87.52 & 91.25 & 98.77 & 93.56 & 94.86 & 92.41 & 98.12 & 81.83 \\
	 & nnPUss & 91.80 & 95.04 & 93.40 & 98.13 & 94.21 & 97.98 & 98.23 & 93.50 & 84.96 \\
	 & $\Delta$ & 0.70 & 7.52 & 2.15 & -0.64 & 0.65 & 3.12 & 5.82 & -4.62 & 3.13 \\
\midrule
	\multirow[c]{3}{*}{0.7} & nnPUcc & 91.64 & 80.18 & 88.83 & 98.84 & 90.62 & 82.96 & 85.41 & 97.50 & 81.63 \\
	 & nnPUss & 94.22 & 96.46 & 94.05 & 99.33 & 95.67 & 99.10 & 98.98 & 96.20 & 87.25 \\
	 & $\Delta$ & 2.58 & 16.27 & 5.22 & 0.49 & 5.06 & 16.14 & 13.57 & -1.30 & 5.62 \\
\midrule
	\multirow[c]{3}{*}{0.9} & nnPUcc & 91.33 & 75.28 & 85.09 & 98.61 & 87.60 & 68.54 & 67.81 & 96.39 & 81.39 \\
	 & nnPUss & 95.70 & 97.57 & 95.15 & 99.80 & 96.94 & 99.43 & 99.19 & 98.95 & 89.39 \\
	 & $\Delta$ & 4.37 & 22.29 & 10.07 & 1.19 & 9.34 & 30.89 & 31.38 & 2.55 & 8.00 \\
\bottomrule
\end{tabular}

    \end{adjustbox}
    \vspace*{3pt} \\
    \begin{adjustbox}{center}
        \begin{tabular}{l|c|ccccccccc}
\toprule
	\textbf{c} & \textbf{Model} & \textbf{California} & \textbf{Credit} & \textbf{Electricity} & \textbf{Wine} & \textbf{20News} & \textbf{HateSpeech} & \textbf{IMDB} & \textbf{PoemSentiment} & \textbf{SMSSpam} \\
\midrule
	\multirow[c]{3}{*}{0.1} & nnPUcc & 81.71 & 64.48 & 75.16 & 69.22 & 79.95 & 88.94 & 73.31 & 84.71 & 88.70 \\
	 & nnPUss & 81.77 & 64.48 & 74.93 & 68.51 & 78.61 & 88.94 & 72.79 & 84.81 & 89.39 \\
	 & $\Delta$ & 0.06 & 0.01 & -0.23 & -0.70 & -1.34 & 0.00 & -0.52 & 0.10 & 0.69 \\
\midrule
	\multirow[c]{3}{*}{0.3} & nnPUcc & 83.20 & 66.59 & 78.42 & 73.89 & 82.50 & 89.52 & 77.00 & 85.00 & 96.55 \\
	 & nnPUss & 84.09 & 67.85 & 78.00 & 74.03 & 81.38 & 89.41 & 75.68 & 85.87 & 95.62 \\
	 & $\Delta$ & 0.89 & 1.26 & -0.42 & 0.14 & -1.12 & -0.10 & -1.32 & 0.87 & -0.92 \\
\midrule
	\multirow[c]{3}{*}{0.5} & nnPUcc & 82.46 & 63.99 & 79.54 & 75.95 & 81.87 & 89.29 & 78.00 & 86.15 & 97.87 \\
	 & nnPUss & 85.31 & 66.23 & 79.64 & 75.97 & 83.28 & 89.32 & 76.83 & 87.60 & 97.18 \\
	 & $\Delta$ & 2.85 & 2.24 & 0.11 & 0.02 & 1.40 & 0.03 & -1.16 & 1.44 & -0.68 \\
\midrule
	\multirow[c]{3}{*}{0.7} & nnPUcc & 80.13 & 62.18 & 78.55 & 76.18 & 78.99 & 89.01 & 77.13 & 89.04 & 98.48 \\
	 & nnPUss & 86.27 & 63.56 & 80.67 & 78.16 & 84.43 & 89.51 & 78.27 & 89.71 & 98.15 \\
	 & $\Delta$ & 6.14 & 1.37 & 2.12 & 1.98 & 5.44 & 0.50 & 1.15 & 0.67 & -0.32 \\
\midrule
	\multirow[c]{3}{*}{0.9} & nnPUcc & 76.82 & 61.33 & 75.87 & 74.72 & 74.27 & 88.82 & 74.76 & 90.00 & 98.72 \\
	 & nnPUss & 86.36 & 60.54 & 81.20 & 79.47 & 85.75 & 89.65 & 79.34 & 90.87 & 98.33 \\
	 & $\Delta$ & 9.54 & -0.79 & 5.33 & 4.76 & 11.48 & 0.83 & 4.58 & 0.87 & -0.39 \\
\bottomrule
\end{tabular}

    \end{adjustbox}
    \label{tab:ss-accuracy}
\end{table}

\begin{table}[htbp]
    \centering
    \scriptsize
    \setlength{\tabcolsep}{3.5pt}
    \renewcommand{\arraystretch}{0.9}
    \caption{Test accuracy, case-control datasets. $\Delta$ indicates accuracy difference between scenario-appropriate $nnPU_{cc}$ method and ill-specified $nnPU_{ss}$ method.}
    \begin{adjustbox}{center}
        \begin{tabular}{l|c|ccccccccc}
\toprule
	\textbf{c} & \textbf{Model} & \textbf{Beans} & \textbf{CIFAR} & \textbf{Chest X-ray} & \textbf{DogFood} & \textbf{EuroSAT} & \textbf{Fashion MNIST} & \textbf{MNIST} & \textbf{Oxford Pets} & \textbf{Snacks} \\
\midrule
	\multirow[c]{3}{*}{0.1} & nnPUss & 77.91 & 93.06 & 89.60 & 85.75 & 86.30 & 94.68 & 93.36 & 82.98 & 73.10 \\
	 & nnPUcc & 78.37 & 93.52 & 88.83 & 86.92 & 89.39 & 96.93 & 95.49 & 86.30 & 72.67 \\
	 & $\Delta$ & 0.47 & 0.46 & -0.77 & 1.16 & 3.09 & 2.25 & 2.13 & 3.31 & -0.43 \\
\midrule
	\multirow[c]{3}{*}{0.3} & nnPUss & 83.88 & 94.40 & 92.99 & 92.43 & 88.26 & 94.00 & 93.50 & 84.24 & 80.40 \\
	 & nnPUcc & 83.64 & 95.48 & 92.15 & 96.44 & 93.26 & 98.47 & 97.56 & 92.56 & 80.77 \\
	 & $\Delta$ & -0.23 & 1.08 & -0.84 & 4.01 & 5.00 & 4.47 & 4.06 & 8.32 & 0.37 \\
\midrule
	\multirow[c]{3}{*}{0.5} & nnPUss & 85.81 & 89.25 & 92.58 & 93.46 & 84.07 & 85.91 & 83.70 & 84.98 & 74.28 \\
	 & nnPUcc & 85.66 & 96.75 & 93.20 & 97.50 & 94.54 & 99.12 & 98.58 & 95.16 & 83.04 \\
	 & $\Delta$ & -0.16 & 7.50 & 0.62 & 4.05 & 10.47 & 13.21 & 14.88 & 10.19 & 8.76 \\
\midrule
	\multirow[c]{3}{*}{0.7} & nnPUss & 81.94 & 81.83 & 82.27 & 88.26 & 76.42 & 83.24 & 85.71 & 79.87 & 69.13 \\
	 & nnPUcc & 85.04 & 97.74 & 90.38 & 96.14 & 95.31 & 99.37 & 99.05 & 96.38 & 80.30 \\
	 & $\Delta$ & 3.10 & 15.91 & 8.11 & 7.88 & 18.89 & 16.13 & 13.33 & 16.51 & 11.18 \\
\midrule
	\multirow[c]{3}{*}{0.9} & nnPUss & 44.34 & 89.43 & 85.44 & 30.24 & 24.56 & 87.21 & 86.61 & 25.70 & 46.79 \\
	 & nnPUcc & 62.40 & 98.71 & 79.86 & 74.77 & 95.55 & 99.58 & 99.25 & 91.36 & 68.05 \\
	 & $\Delta$ & 18.06 & 9.28 & -5.57 & 44.53 & 70.99 & 12.37 & 12.64 & 65.67 & 21.26 \\
\bottomrule
\end{tabular}

    \end{adjustbox}
    \vspace*{3pt} \\
    \begin{adjustbox}{center}
        \begin{tabular}{l|c|ccccccccc}
\toprule
	\textbf{c} & \textbf{Model} & \textbf{California} & \textbf{Credit} & \textbf{Electricity} & \textbf{Wine} & \textbf{20News} & \textbf{HateSpeech} & \textbf{IMDB} & \textbf{PoemSentiment} & \textbf{SMSSpam} \\
\midrule
	\multirow[c]{3}{*}{0.1} & nnPUss & 81.27 & 63.93 & 74.60 & 68.42 & 78.46 & 88.00 & 72.55 & 82.57 & 89.27 \\
	 & nnPUcc & 81.27 & 64.07 & 74.82 & 69.61 & 79.82 & 88.00 & 73.14 & 82.76 & 88.68 \\
	 & $\Delta$ & 0.00 & 0.14 & 0.22 & 1.19 & 1.36 & 0.00 & 0.59 & 0.19 & -0.58 \\
\midrule
	\multirow[c]{3}{*}{0.3} & nnPUss & 83.74 & 67.59 & 76.33 & 69.96 & 79.63 & 86.40 & 71.35 & 82.86 & 94.20 \\
	 & nnPUcc & 84.60 & 67.89 & 76.95 & 73.55 & 83.53 & 86.44 & 75.72 & 82.86 & 95.64 \\
	 & $\Delta$ & 0.86 & 0.30 & 0.62 & 3.59 & 3.90 & 0.05 & 4.36 & 0.00 & 1.43 \\
\midrule
	\multirow[c]{3}{*}{0.5} & nnPUss & 84.03 & 69.67 & 77.08 & 72.15 & 73.08 & 83.00 & 71.01 & 82.10 & 92.26 \\
	 & nnPUcc & 85.78 & 70.68 & 77.51 & 76.46 & 85.86 & 84.28 & 76.32 & 82.38 & 96.83 \\
	 & $\Delta$ & 1.75 & 1.01 & 0.43 & 4.32 & 12.78 & 1.28 & 5.31 & 0.29 & 4.57 \\
\midrule
	\multirow[c]{3}{*}{0.7} & nnPUss & 85.45 & 71.69 & 77.47 & 72.11 & 72.80 & 71.71 & 70.83 & 72.38 & 77.40 \\
	 & nnPUcc & 87.44 & 73.94 & 78.20 & 78.63 & 88.64 & 80.01 & 77.38 & 79.90 & 96.48 \\
	 & $\Delta$ & 1.99 & 2.25 & 0.73 & 6.52 & 15.84 & 8.31 & 6.55 & 7.52 & 19.08 \\
\midrule
	\multirow[c]{3}{*}{0.9} & nnPUss & 83.13 & 71.34 & 76.06 & 70.78 & 81.35 & 44.69 & 70.96 & 34.57 & 40.71 \\
	 & nnPUcc & 89.85 & 76.13 & 79.58 & 82.97 & 92.80 & 63.64 & 79.70 & 59.14 & 89.85 \\
	 & $\Delta$ & 6.72 & 4.79 & 3.51 & 12.19 & 11.46 & 18.95 & 8.74 & 24.57 & 49.14 \\
\bottomrule
\end{tabular}

    \end{adjustbox}
    \label{tab:cc-accuracy}
\end{table}

Experiment results are summarized in Tables~\ref{tab:ss-accuracy} and~\ref{tab:cc-accuracy}, for s-s and c-c data respectively (results for metrics other than accuracy: precision, recall and F1 score can be found in the GitHub repository; observations for F1 score largely correlate with the accuracy-based analysis below). The key observation is that in both cases the advantage of the correctly specified method (defined as the difference of respective accuracies and denoted by $\Delta$) starts for $c$   as low $c = 0.1$, but as the proportion of labeled examples in the dataset increases it tends to significantly increase. Upon closer inspection, this is expected -- as apparent in Figure~\ref{fig:ss_vs_cc}, the difference in unlabeled sample structure is subtle when label frequency is low, but becomes very apparent  when $c$ increases. That causes both methods' performance to be close when label frequency is low, but for high $c$ values, when scenario dependency deepens, they start to diverge.

\begin{figure}[tb]
\vspace*{-1mm}
    \centering
    \includegraphics[width=\textwidth]{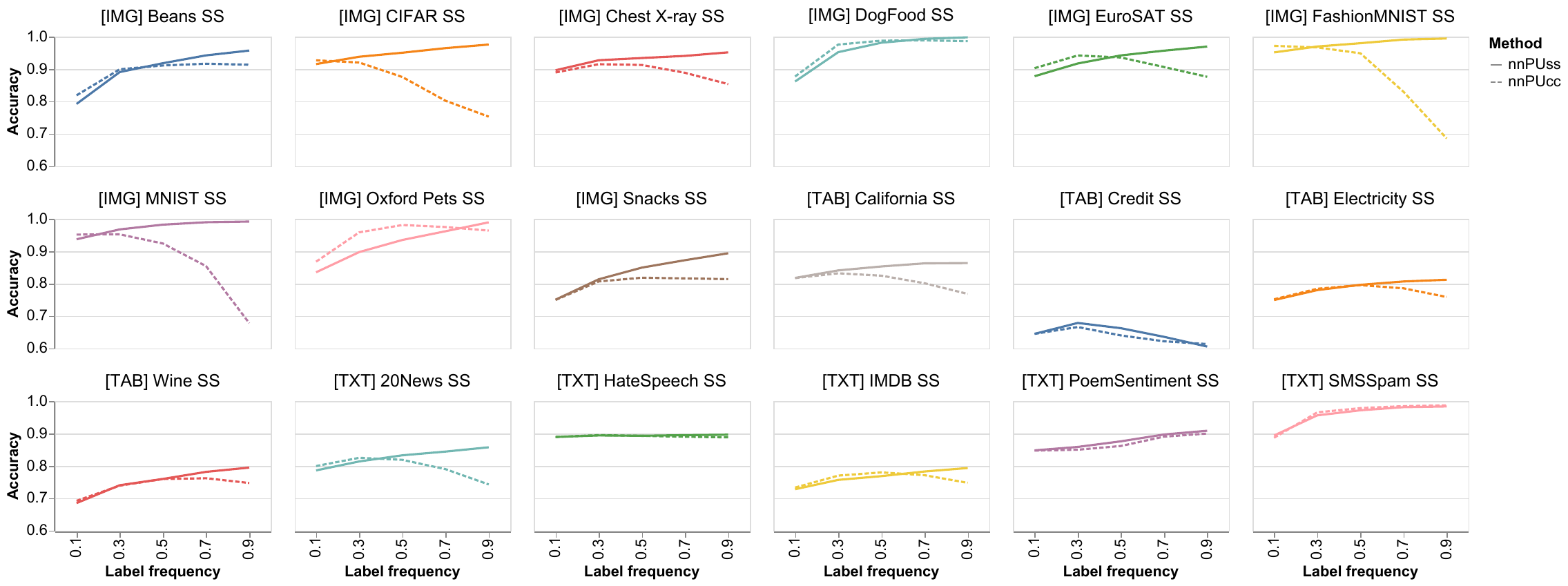}\vspace*{-1mm}
    \caption{Change of accuracy with label frequency increase for single-sample datasets}
    \label{fig:lf_increase_ss}
\end{figure}

\begin{figure}[tb]
    \centering
    \includegraphics[width=\textwidth]{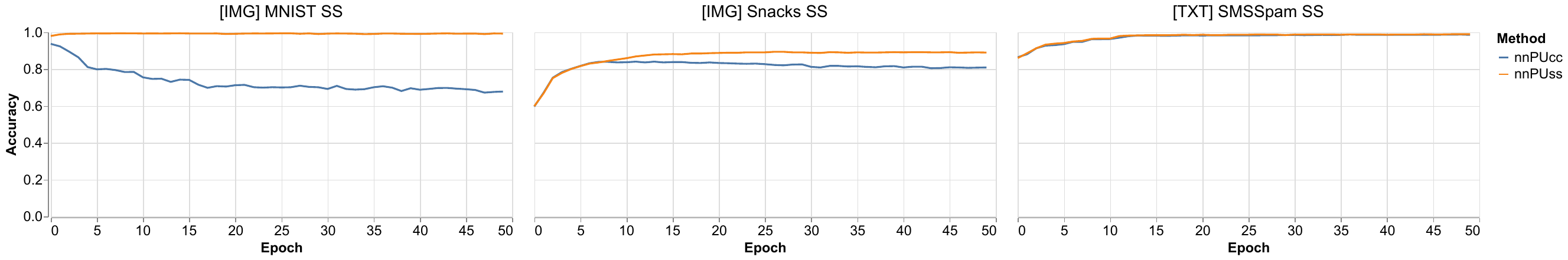}\vspace*{-1mm}
    \caption{Test accuracy per epoch, selected single-sample datasets, $c = 0.9$}
    \label{fig:overfitting}\vspace*{-2mm}
\end{figure}

\medskip
Due to the reasons above, it is worth inspecting results for $c = 0.9$ in detail, as this is when scenario differences are the most distinctive. In the vast majority of cases, the advantage of correctly specified method is very significant -- this is most apparent for some image datasets, such as MNIST and CIFAR. Rare cases where performance of both methods is similar are more common for text and tabular data. Note that the for the correct method, in vast majority of cases accuracy tends to steadily increase as the label frequency rises, while the alternative methods' performance starts to drop off. Figure~\ref{fig:lf_increase_ss} illustrates this phenomenon for s-s datasets, but the former statement holds true for case-control data as well. Depicted behavior seems to indicate overfitting of the ill-specified method for high $c$ values, which, as apparent in Figure~\ref{fig:overfitting}, occurs on multiple datasets and is the major cause of the classification performance deterioration. We stress that as  in the positive-unlabeled problems we have no access to the fully labeled validation dataset, robustness to overfitting  is drastically  more important for any PU learning method than in the binary classification task. Overall, ill-specified methods are prone to overfitting after only a couple of epochs. Rare cases, where they do not overfit, match cases, where the overall accuracy of both methods stay close.

\begin{figure}[tbp]
    \centering
    \includegraphics[width=\textwidth]{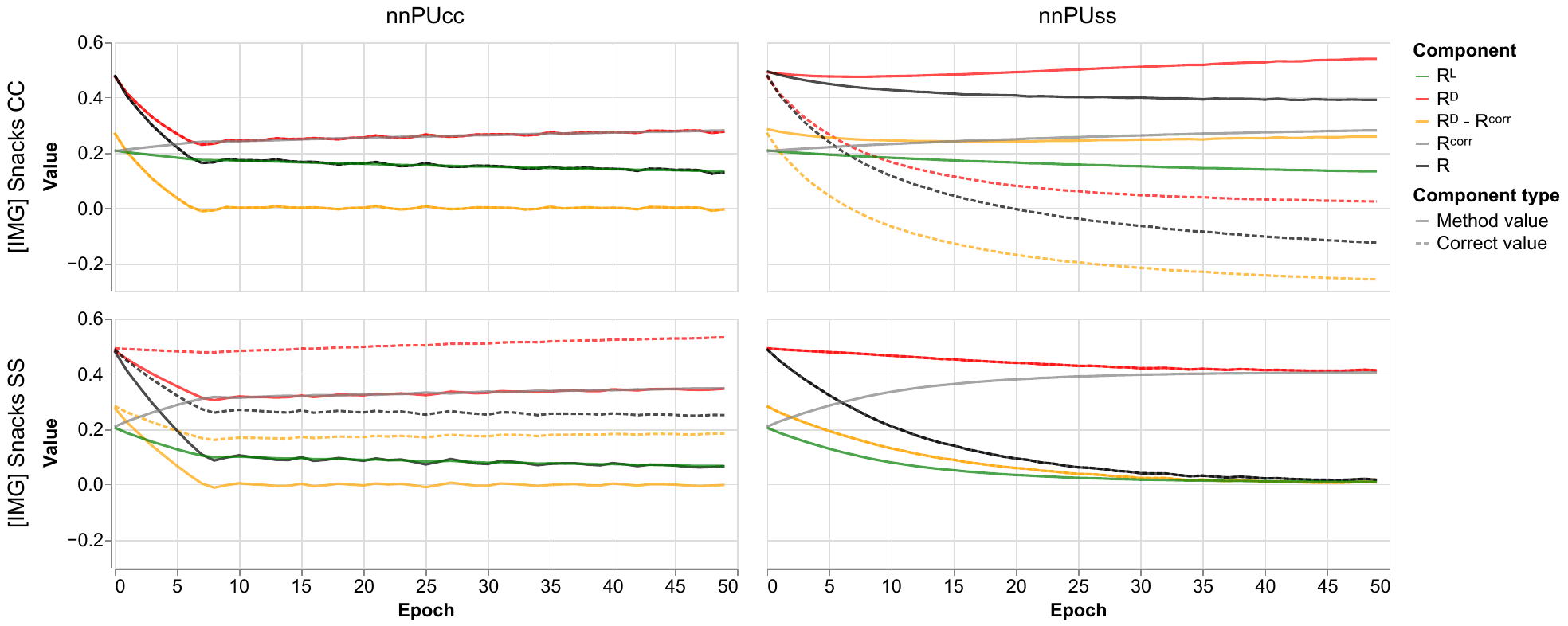}
    \caption{Risk components per epoch, Snacks dataset, $c = 0.9$. ,,Method'' values refer to risk values obtained during training, whereas ,,Correct'' values -- to the ones which would be obtained in the given epoch if scenario-aware risk would be applied.}
    \label{fig:loss_components}
\end{figure}

\eject
In order to understand the behavior of both methods better, closer inspection of a change of risk values during training is crucial. Figure~\ref{fig:loss_components} illustrates a typical risk component changes throughout the learning process for both $nnPU_{ss}$ and $nnPU_{cc}$. Note that even when wrong method is applied ($nnPU_{cc}$ for single-sample datasets and vice versa) the ,,correct'' risk values still decrease throughout training. This suggests that for both $nnPU_{ss}$ and $nnPU_{cc}$ gradients might point in the similar directions, and explains why (especially in the initial training epochs) learning does not fail completely. It is also interesting to consider reasons why this does not hold true throughout whole training. When attempting to train $nnPU_{ss}$ on the case-control data (Fig.~\ref{fig:loss_components}, upper right), note that  for the correct, case-control-tailored risk value, non-negative risk component $R^D - R^{corr}$ drops well below 0 -- which is strongly undesirable  and is normally discouraged in Algorithm~\ref{algorithm:nnPU-upgrade}, step~\ref{algorithm:nnPU-upgrade:negative-correction}. Due to this, severe overfitting occurs only a few epochs into training. In the symmetric case, $nnPU_{cc}$ on the single-sample data (Fig.~\ref{fig:loss_components}, lower left), the problem is slightly different. Due to $R^D$ being underestimated by the algorithm (as it expect more positive samples in the unlabeled dataset), nonnegative component of the risk quickly falls towards 0 and starts oscillating in its vicinity. This time step~\ref{algorithm:nnPU-upgrade:negative-correction} of algorithm~\ref{algorithm:nnPU-upgrade} is activated  very early. However, by the view of the correct, scenario-aware risk function, these updates are counterproductive -- they are only valid when nonnegative component stops being positive -- and counteract quasi-valid updates of the model, causing performance decline.

\section{Conclusions}

In this paper,  we  discussed  issues with automatic application of PU learning methods without taking into account labeling scenario constraints. To this end, we introduced $nnPU_{ss}$ -- a tailored  to s-s scenario equivalent of $nnPU$ (denoted in this paper as $nnPU_{cc}$), a popular benchmark method in PU learning papers -- and through experiments, we exposed dangers of incorrect method selection, and identified reasons for such behavior. Obtained results indicate that for high-enough label frequencies (approximately $c \geq 0.5$), using algorithms not devised  for a particular scenario might incur a major performance loss, and -- especially in the case of ranking  classifiers according to their performance -- might put the method investigated  at the huge disadvantage. We also introduced a correct form of $nnPU_{ss}$ classifier for s-s data and indicated that its algorithm can be easily obtained from the algorithm  of $nnPU_{ss}$ classifier by changing one line of its code.
Future research on this topic might include expanding this analysis to more  types of methods, as well as identify  dependence  of this phenomenon on methods' complexity and/or its training parameters.  It would be  also desirable to analyze which of the methods devised for c-c scenario can be easily adapted, as it is the case for ERM classifiers, to s-s set up.

\end{document}